\documentclass[10pt,twocolumn,letterpaper]{article}

\usepackage{cvpr}
\usepackage{times}
\usepackage{epsfig}
\usepackage{graphicx}
\usepackage{amsmath}
\usepackage{amssymb}
\usepackage{bbm}
\usepackage{array}
\usepackage{multirow}
\usepackage{xcolor}
\usepackage[perpage]{footmisc}
\newcommand{\PreserveBackslash}[1]{\let\temp=\\#1\let\\=\temp}
\newcolumntype{C}[1]{>{\PreserveBackslash\centering}p{#1}}
\newcolumntype{R}[1]{>{\PreserveBackslash\raggedleft}p{#1}}
\newcolumntype{L}[1]{>{\PreserveBackslash\raggedright}p{#1}}

\usepackage[pagebackref=true,breaklinks=true,letterpaper=true,colorlinks,bookmarks=false]{hyperref}

\cvprfinalcopy 

\newif\ifsti
\stitrue

\newif\iflyft
\lyfttrue

\def\cvprPaperID{956} 

\ifcvprfinal\fi
\begin{document}

\ifsti
\title{STINet: Spatio-Temporal-Interactive Network for Pedestrian Detection and Trajectory Prediction}
\fi

\newcommand{\apmb}[3][\normalsize]{{#1{#2}}{\tiny$\pm$#3}}
\newcommand{\apmbbf}[3][\normalsize]{\textbf{{#1{#2}}{\tiny$\pm$#3}}}
\newcommand{\newstuff}[1]{\textcolor{red}{#1}}

\author{Zhishuai Zhang$^{1,2*}$ \quad Jiyang Gao$^{1}$ \quad Junhua Mao$^1$  \quad Yukai Liu$^1$ \quad Dragomir Anguelov$^1$ \quad Congcong Li$^1$ \\
$^1$Waymo LLC \qquad $^2$ Johns Hopkins University \\
{\tt\small zzhang99@jhu.edu, \quad\{jiyanggao, junhuamao, liuyukai, dragomir, congcongli\}@waymo.com}}

\maketitle
\thispagestyle{empty}
\newcommand\blfootnote[1]{%
  \begingroup
  \renewcommand\thefootnote{}\footnote{#1}%
  \addtocounter{footnote}{-1}%
  \endgroup
}

\blfootnote{$*$ Work done during an internship at Waymo.}
\begin{abstract}
Detecting pedestrians and predicting future trajectories for them are critical tasks for numerous applications, such as autonomous driving.
Previous methods either treat the detection and prediction as separate tasks or simply add a trajectory regression head on top of a detector.
\ifsti
In this work, we present a novel end-to-end two-stage network: Spatio-Temporal-Interactive Network (STINet).
\fi
In addition to 3D geometry modeling of pedestrians, we model the temporal information for each of the pedestrians. To do so, our method predicts both current and past locations in the first stage, so that each pedestrian can be linked across frames and the comprehensive spatio-temporal information can be captured in the second stage. Also, we model the interaction among objects with an interaction graph, to gather the information among the neighboring objects.
Comprehensive experiments on the Lyft Dataset and the recently released large-scale Waymo Open Dataset for both object detection and future trajectory prediction validate the effectiveness of the proposed method. For the Waymo Open Dataset, we achieve a bird-eyes-view (BEV) detection AP of 80.73 and trajectory prediction average displacement error (ADE) of 33.67cm for pedestrians, which establish the state-of-the-art for both tasks.
\end{abstract}

\section{Introduction}

\begin{figure}
    \centering
    \includegraphics[width=\linewidth]{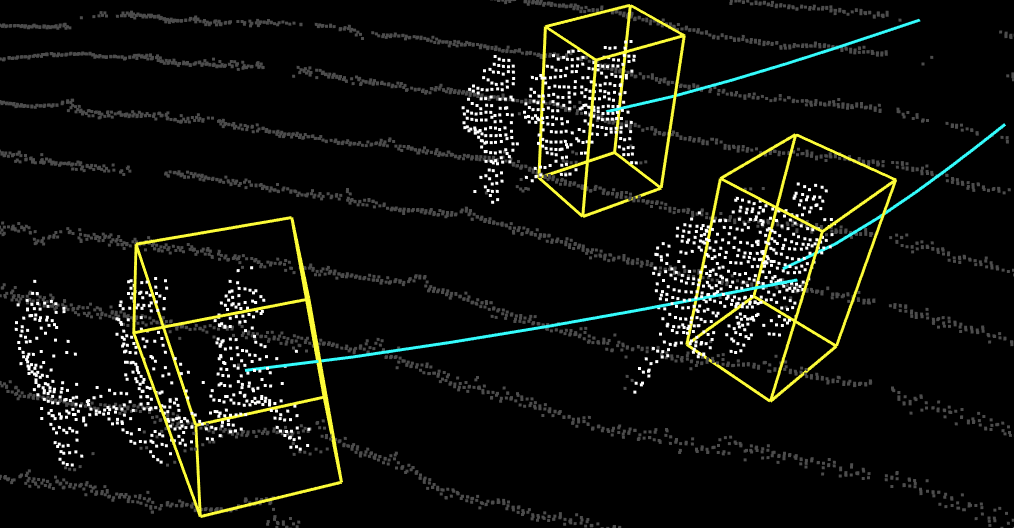}
    \caption{Given a sequence of current and past point clouds, our task is to detect pedestrians in the current frame, and predict the future trajectory of them. In this figure, white points are input point cloud sequence (stacked for visualization), yellow boxes are detected objects, and the cyan lines are predicted future trajectory.}
    \label{fig:task}
\end{figure}

To drive safely and smoothly, self-driving cars (SDC) not only need to detect where the objects are currently (\ie object detection), but also need to predict where they will go in the future (\ie trajectory prediction). Among the objects, pedestrian is an important and difficult type. The difficulty comes from the complicated properties of pedestrian appearance and behavior, \eg deformable shape and interpersonal relations~\cite{gupta2018social}. In this paper, we tackle the problem of joint pedestrian detection and trajectory prediction from a sequence of point clouds, as illustrated in Figure~\ref{fig:task}.

Traditionally, this problem is tackled by dividing the perception pipeline into multiple modules: object detection~\cite{girshick2015fast,lang2019pointpillars,lin2017focal,liu2016ssd,redmon2016you,ren2015faster,zhou2019end,zhou2018voxelnet}, tracking~\cite{milan2017online} and trajectory prediction~\cite{alahi2016social,gupta2018social,hong2019rules}; latter modules take the outputs from the former modules. Although such strategy makes each sub-module easy to design and implement, it sacrifices the potential advantage of joint optimization. Latter modules can lose critical information bottle-necked by the interfaces between sub-modules, \eg a pedestrian's future trajectory depends on many useful geometry features from the raw sensor data, which may be abstracted away in the detection/tracking stage. To this end, researchers recently have proposed several end-to-end neural networks to detect objects and predict trajectories simultaneously. FaF~\cite{luo2018fast} and IntentNet~\cite{casas2018intentnet} are two of the representative methods, which are designed based on single stage detectors (SSD)~\cite {liu2016ssd};
in addition to original anchor classification and regression of SSD, they also regress a future trajectory for each anchor.

We observed that there are two major issues that are critical for joint detection and trajectory prediction, but are not addressed by previous end-to-end methods: 1) Temporal modeling on object level: existence and future trajectory of an object are embedded in both current and past frames. Current methods simply reuse single-stage detector and fuse the temporal information in the backbone CNN in an object-agnostic manner either via feature concatenation or 3D CNN~\cite{casas2018intentnet,luo2018fast}. Such coarse level fusion can loss fine-grained temporal information for each object, which is critical for both tasks.
2) Interaction modeling among objects: the future trajectory of an object could be influenced by the other objects. \Eg, a pedestrian walking inside a group may tend to follow others. Existing methods~\cite{casas2018intentnet,luo2018fast} do not explicitly model interactions among objects.

To address the aforementioned issues, we propose an end-to-end \ifsti Spatio-Temporal-Interactive network (STINet) \fi to model pedestrians’ temporal and interactive information jointly. The proposed network takes a sequence of point clouds as input, detects current location and predicts future trajectory for pedestrians. Specifically, there are three sub-components in \ifsti STINet \fi: backbone network, proposal generation network, and proposal prediction network. In the backbone net, we adopted a similar structure as PointPillars~\cite{lang2019pointpillars}, and applied it on each frame of the point cloud, the output feature maps from multi-frames are then combined. The proposal generation network takes feature maps from the backbone net and generates potential pedestrian instances with both their current and past locations (\ie temporal proposals); such temporal proposals allow us to link the same object across different frames. In the third module (\ie prediction network), we use the temporal proposals to explicitly gather the geometry appearance and temporal dynamics for each object. To reason the interaction among pedestrians, we build a graph layer to gather the information from surrounding pedestrians. After extracting the above spatial-temporal-interactive feature for each proposal, the detection and prediction head uses the feature to regress current detection bounding box and future trajectory.

Comprehensive experiments are conducted on Waymo Open Dataset~\cite{waymo2019open} \iflyft and Lyft Dataset~\cite{lyft2019} \fi to demonstrate the effectiveness of the \ifsti STINet\fi. Specifically, it achieves an average precision of 80.73 for bird-eyes-view pedestrian detection, and an average displacement error of 33.67 cm for trajectory prediction \iflyft on Waymo Open Dataset\fi. It achieves real-time inference speeds and takes only 74.6 ms for inference on a range of 100m by 100m.

The main contributions of our work come in four folds:
\begin{itemize}
	\item We build an end-to-end network tailored to model pedestrian past, current and future simultaneously.
	\item We propose to generate temporal proposals with both current and past boxes. This enables learning a comprehensive spatio-temporal representation for pedestrians with their geometry, dynamic movement and history path in an end-to-end manner without explicitly associating object across frames.
	\item We propose to build a graph among pedestrians to reason the interactions to further improve trajectory prediction quality.
	\item We establish the state-of-the-art performance for both detection and trajectory prediction on the Lyft Dataset and the recent large-scale challenging Waymo Open Dataset.
\end{itemize}


\section{Related work}
\subsection{Object detection}
Object detection is a fundamental task in computer vision and autonomous driving. Recent approaches can be divided into two folds: single-stage detection~\cite{lin2017focal,liu2016ssd,redmon2016you} and two-stage detection~\cite{girshick2015fast,ren2015faster}. Single-stage detectors do classification and regression directly on backbone features, while two-stage detectors generate proposals based on backbone features, and extract proposal features for second-stage classification and regression. Single-stage detectors have simpler structure and faster speed, however, they lose the possibility to flexibly deal with complex objects behaviors, \eg, explicitly capturing pedestrians moving across frames with different speeds and history paths. In this work, we follow the two-stage detection framework and predict object boxes for both current and past frames as proposals, which are further processed to extract their geometry and movement features.
\begin{figure*}
    \centering
    \includegraphics[width=0.95\linewidth]{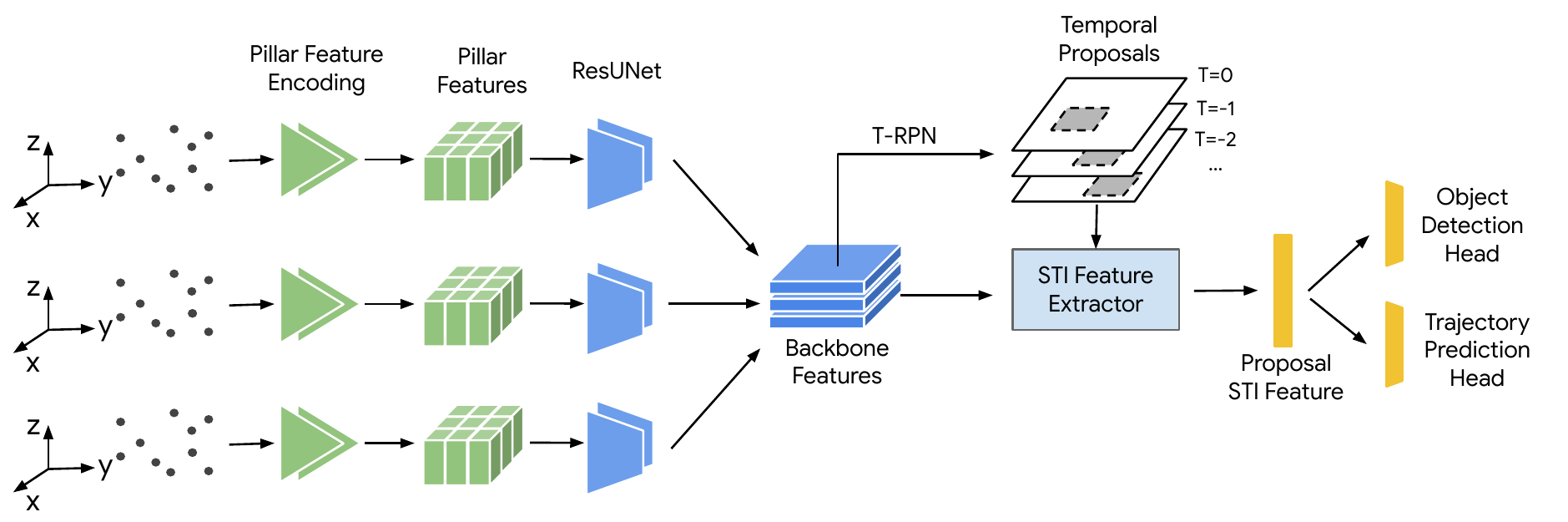}
    \caption{The overview of the proposed method. It takes a sequence of point clouds as input, detects pedestrians and predicts their future trajectories simultaneously. The point clouds are processed by Pillar Feature Encoding~\cite{lang2019pointpillars,zhou2018voxelnet} to generate Pillar Features. Then each Pillar Feature is fed into a backbone ResUNet~\cite{ronneberger2015u} to get backbone features. A Temporal Region Proposal Network (T-RPN) takes backbone features and generated temporal proposal with past and current boxes for each object. \ifsti Spatio-Temporal-Interactive (STI) \fi Feature Extractor learns features for each temporal proposal which are used for final detection and trajectory prediction.}
    \label{fig:overall_pipeline}
\end{figure*}

\subsection{Temporal proposals}
Temporal proposals have been shown beneficial in action localization in~\cite{hou2017tube,kalogeiton2017action}. They showed associating temporal proposals from different video clips can help to leverage the temporal continuity of video frames. \cite{tang2019object} proposed to link temporal proposals throughout the video to improve video object detection. In our work, we also exploit temporal proposals and step further to investigate and propose how to build comprehensive spatio-temporal representations of proposals to improve future trajectory prediction. This is a hard task since there are no inputs available for the future. Also we investigate to learn interactions between proposals via a graph. We show that these spatio-temporal features can effectively model objects' dynamics and provide accurate detection and prediction of their future trajectory.

\subsection{Relational reasoning}
An agent's behavior could be influenced by other agents and it is naturally connected to relational reasoning~\cite{battaglia2018relational,santoro2017simple}. Graph neural networks have shown its strong capability in relational modeling in recent years. Wang \etal formulated the video as a space-time graph, show the effectiveness on the video classification task~\cite{wang2018videos}. Sun \etal designed a relational recurrent network for action detection and anticipation~\cite{sun2019relational}. Yang \etal proposed to build an object relationship graph for the task of scene graph generation~\cite{yang2018graph}.

\subsection{Trajectory prediction}
Predicting the future trajectory of objects is an important task, especially for autonomous driving. Previous research has been conducted based on perception objects as inputs~\cite{alahi2016social,chang2019argoverse,gupta2018social,hong2019rules,lee2017desire}. Recently FaF~\cite{luo2018fast} and IntentNet~\cite{casas2018intentnet} focused on end-to-end trajectory prediction from raw point clouds as input. However, they simply re-used single-stage detection framework and added new regression heads on it. In our work, we exploit temporal region proposal network and explicitly model \ifsti Spatio-Temporal-Interaction (STI) \fi representations of pedestrians, and our experiments show that the proposed \ifsti STI \fi modeling is superior on both detection and trajectory prediction for pedestrians.
\begin{figure}
    \centering
    \includegraphics[width=0.95\linewidth]{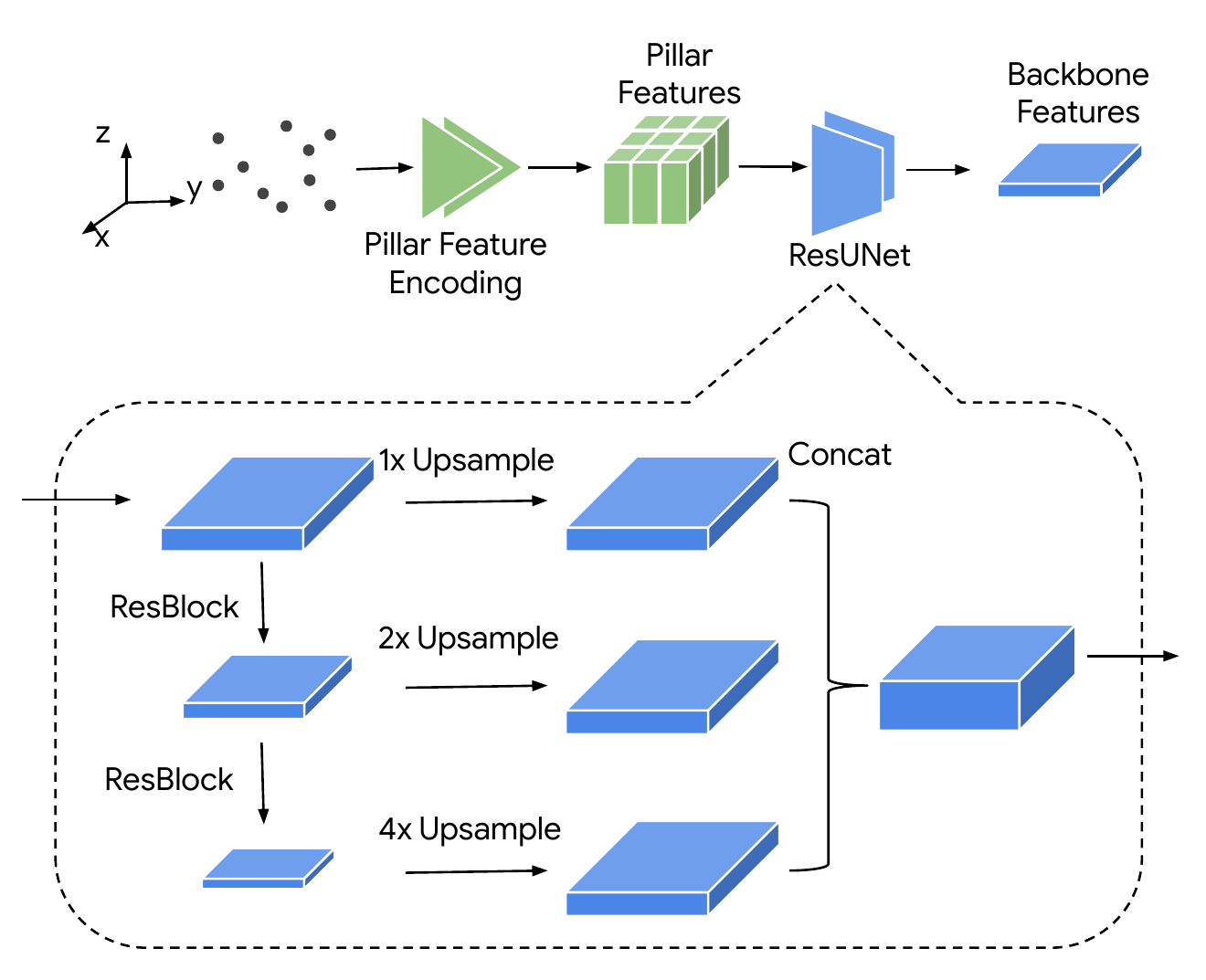}
    \caption{Backbone of proposed network. Upper: overview of the backbone. The input point cloud sequence is fed to Voxelization and Point net to generate pseudo images, which are then processed by ResNet U-Net to generate final backbone feature sequence. Lower: detailed design of ResNet U-Net.}
    \label{fig:backbone}
\end{figure}

\section{Proposed method}
In this section, we discuss our proposed network in details. The overview of our proposed method is shown in Figure~\ref{fig:overall_pipeline}, which can be divided into three steps. For each of these steps, we discuss in the following subsections.
\subsection{Backbone network}\label{sec:backbone}
The backbone of our network is illustrated in Figure~\ref{fig:backbone}. The input is a sequence of point clouds with $t'$ frames noted as $[\text{PC}_{-(t'-1)}, \text{PC}_{-(t'-2)},\cdots,\text{PC}_{0}]$, which corresponds to the lidar sensor input from the past $t'-1$ frames as well as the current frame. All point clouds are calibrated to SDC’s pose at the current frame so that the ego-motion is discarded. To build rich pillar features while keeping a feasible memory usage, we generate $t$ pillar features from the $t'$ input frames. Consecutive $t'/t$ point clouds $\text{PC}_{-(j+1)t'/t+1},\cdots,\text{PC}_{-jt'/t}$ are processed with Voxelization~\cite{lang2019pointpillars,zhou2018voxelnet} and then concatenated to generate a pseudo image $I_j$ (\ie Pillar Features) with shape $H\times W\times C_{in}$. Thus the output of Pillar Feature Encoding is a sequence of $t$ Pillar Features $[I_{-(t-1)},I_{-(t-2)},\cdots,I_{0}]$.

Next we adopt a similar backbone CNN network proposed as in~\cite{ronneberger2015u}, as shown in the lower part of Figure~\ref{fig:backbone}. Each of the Pillar Features $I_j$ is first processed by three ResNet-style blocks to generate intermediate features with shape $\mathbb{R}^{H\times W\times C_{0}}$,$\mathbb{R}^{\frac{1}{2}H\times\frac{1}{2}W\times C_{1}}$ and $\mathbb{R}^{\frac{1}{4}H\times\frac{1}{4}W\times C_{2}}$. Then we use deconvolution layers to upsample them to the same spatial shape with $I_j$. The concatenation of the upsampled features serve as the backbone feature of $I_j$, noted as $B_j$.

\subsection{Temporal proposal generation}\label{sec:trpn}
In order to explicitly model objects' current and past knowledge, we propose a temporal region proposal network (T-RPN) to generate object proposals with both current and past boxes. T-RPN takes the backbone feature sequence $[B_{-(t-1)},B_{-(t-2)},\cdots,B_{0}]$ as the input, concatenates them in the channel dimension and applies a $1\times 1$ convolution to generate a temporal-aware feature map. Classification, current frame regression and past frames regression are generated by applying $1\times 1$ convolutional layers over the temporal-aware feature map, to classify and regress the pre-defined anchors.

The temporal region proposal network is supervised by ground-truth objects' current and past locations. For each anchor $\mathbf{a}=(x^a, y^a, w^a, l^a, h^a)$ ($x$, $y$, $w$, $l$, $h$ correspond to x coordinate of box center, y coordinate of box center, width of box, length of box and heading of box respectively), it is assigned to a ground-truth object with largest IoU of the current frame box $\mathbf{gt}=(x_0^{gt}, y_0^{gt}, w^{gt}, l^{gt}, h_0^{gt})$. Similar to SECOND~\cite{yan2018second}, we compute the regression target in order to learn the difference between the pre-defined anchors and the corresponding ground-truth boxes. For the current frame, we generate a 5-d regression target $\mathbf{d}^a_0=(dx^a_0,dy^a_0,dw^a,dl^a,dh^a_0)$:
\begin{equation}
    dx^a_0=(x^{gt}_0-x^a)/\sqrt{(x^a)^2+(y^a)^2}\label{eq:1}
\end{equation}
\begin{equation}
    dy^a_0=(y^{gt}_0-y^a)/\sqrt{(x^a)^2+(y^a)^2}
\end{equation}
\begin{equation}
    dw^a=\log{\frac{w^{gt}}{w^a}}
\end{equation}
\begin{equation}
    dl^a=\log{\frac{l^{gt}}{l^a}}
\end{equation}
\begin{equation}
    dh^a_0=\sin{\frac{h^{gt}_0-h^a}{2}}\label{eq:5}
\end{equation}
With similar equations, we also compute $t-1$ past regression targets for anchor $\mathbf{a}$ against the same ground-truth object: $\mathbf{d}^a_j=(dx^a_j,dy^a_j,dh^a_j)$ for $j\in\{-1,-2,\cdots,-(t-1)\}$. Width and length are not considered for the past regression since we assume the object size does not change across different frames. For each anchor $\mathbf{a}$, the classification target $s^a$ is assigned as $1$ if the assigned ground-truth object has an IoU greater than $th^+$ at the current frame. If the IoU is smaller than $th^-$, classification target is assigned as $0$. Otherwise the classification target is $-1$ and the anchor is ignored for computing loss.

For each anchor $\mathbf{a}$, T-RPN predicts a classification score $\hat{s}^a$, a current regression vector $\hat{\mathbf{d}}^a_0=(\hat{d}x^a_0,\hat{d}y^a_0,\hat{d}w^a,\hat{d}l^a,\\\hat{d}h^a_0)$ and $t-1$ past regression vectors $\hat{\mathbf{d}}^a_j=(\hat{d}x^a_j,\hat{d}y^a_j,\\\hat{d}h^a_j)$ from the aforementioned $1\times 1$ convolutional layers. The objective of T-RPN is the weighted sum of classification loss, current frame regression loss and past frames regression loss as defined in the equations below, where $\mathbbm{1}(x)$ is the indicator function and returns $1$ if $x$ is true otherwise $0$.
\begin{equation}
    \mathcal{L}_{\text{T-RPN}}=\lambda_{\text{cls}}\mathcal{L}_{\text{cls}}+\lambda_{\text{cur\_reg}}\mathcal{L}_{\text{cur\_reg}}+\lambda_{\text{past\_reg}}\mathcal{L}_{\text{past\_reg}}\label{eq:6}
\end{equation}
\begin{equation}
    \mathcal{L}_{\text{cls}}=\frac{\sum_{\mathbf{a}}\text{CrossEntropy}(s^a,\hat{s}^a)\mathbbm{1}(s^a\ge 0)}{\sum_{\mathbf{a}}\mathbbm{1}(s^a\ge 0)}
\end{equation}
\begin{equation}
    \mathcal{L}_{\text{cur\_reg}}=\frac{\sum_{\mathbf{a}}\text{SmoothL1}(\mathbf{d}^a_0,\hat{\mathbf{d}}^a_0)\mathbbm{1}(s^a\ge 1)}{\sum_{\mathbf{a}}\mathbbm{1}(s^a\ge 1)}
\end{equation}
\begin{equation}
    \mathcal{L}_{\text{past\_reg}}=\sum_{j=1}^{t-1}\frac{\sum_{\mathbf{a}}\text{SmoothL1}(\mathbf{d}^a_{-j},\hat{\mathbf{d}}^a_{-j})\mathbbm{1}(s^a\ge 1)}{\sum_{\mathbf{a}}\mathbbm{1}(s^a\ge 1)}\label{eq:9}
\end{equation}

For proposal generation, classification scores and regression vectors are applied on pre-defined anchors to generate temporal proposals, by reversing Equations~\ref{eq:1}-\ref{eq:5}. Thus each temporal proposal has a confidence score as well as the regressed boxes for the current and past frames. After that, non-maximum suppression is applied on the current frame boxes of temporal proposals to remove redundancy.

\subsection{Proposal prediction}
\subsubsection{\ifsti Spatio-temporal-interactive \fi feature extraction}\label{sec:proposal_feature_extraction}

Given backbone features $[B_{-(t-1)}, \cdots, B_0]$ and temporal proposals, spatio-temporal-interactive features are learned for each temporal proposal to capture the comprehensive information for detection and trajectory prediction. Different ways for modeling objects are combined to achieve this.

\begin{figure}
    \centering
    \includegraphics[width=\linewidth]{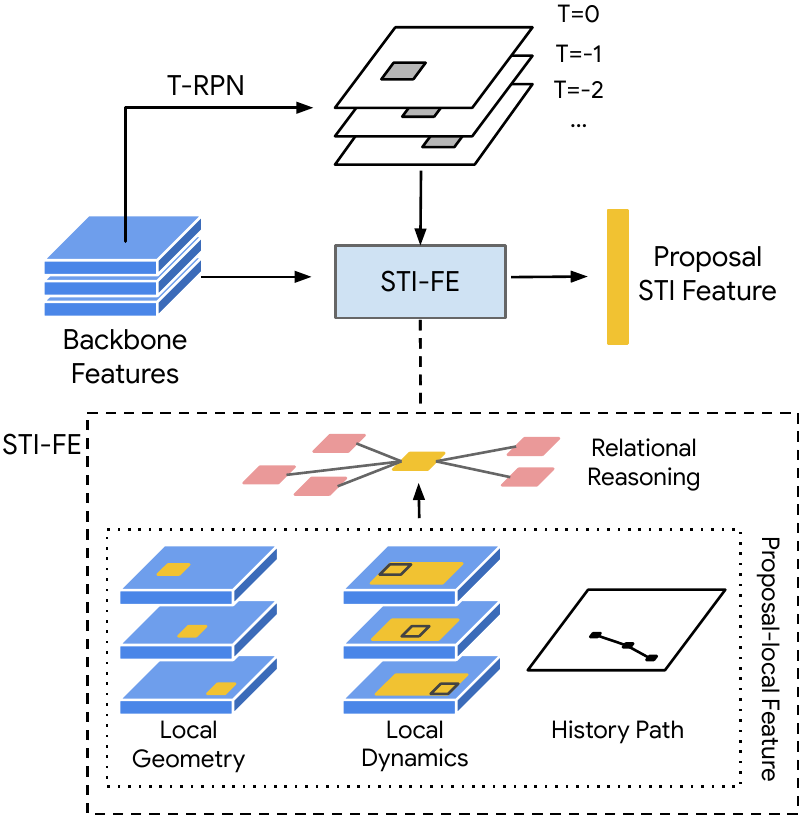}
    \caption{Spatial-Temporal-Interactive Feature Extractor (STI-FE): Local geometry, local dynamic and history path features are extracted given a temporal proposal. For local geometry and local dynamics features, the yellow areas are used for feature extraction. Relational reasoning is performed across proposals' local features to generate interactive features.}
    \label{fig:feature_extraction}
\end{figure}

\noindent
\textbf{Local geometry feature:}
To extract object geometry knowledge, we use the proposal boxes at j-th frame (\ie $x_j$, $y_j$, $w$, $l$, and $h_j$) to crop features from $B_j$, as shown in the lower left part of Figure~\ref{fig:feature_extraction}. This is an extension of traditional proposal feature cropping used in Faster-RCNN~\cite{ren2015faster}, to gather position-discarded local geometry features from each frame. To simplify the implementation on TPU, we rotate the 5-DoF box $(x_j,y_j,w,l,h_j)$ to the closest standing box $(x_{min,j},y_{min,j},x_{max,j},y_{max,j})$ for ROIAlign~\cite{he2017mask}.

\noindent
\textbf{Local dynamic feature:}
As illustrated in the lower middle part of Figure~\ref{fig:feature_extraction}, we use a meta box (drawn in yellow) which covers the whole movement of the pedestrian to crop features for all $B_j$'s. The meta box is the smallest box which contains all current and history proposal boxes. Formally, after transferring all rotated proposal boxes $(x_j,y_j,w,l,h_j)$ to the closest standing boxes $(x_{min,j},y_{min,j},x_{max,j},y_{max,j})$, the meta box is computed with the following equations:
\begin{equation*}
x_{min}=\min_j(x_{min,j});y_{min}=\min_j(y_{min,j})
\end{equation*}
\begin{equation*}
x_{max}=\max_j(x_{max,j});y_{max}=\max_j(y_{max,j})
\end{equation*}
This feature captures the direction, curvature and speed of the object, which are useful for future trajectory prediction.

\noindent
\textbf{History path feature:}
In order to directly encode objects' past movement, we exploit the location displacement over different frames as the history path feature. To be specific, given a temporal proposal with $x_j$, $y_j$ as the box centers, the history path feature is $\text{MLP}([x_0-x_{-1},y_0-y_{-1},x_0-x_{-2},y_0-y_{-2},\cdots,x_0-x_{-(t-1)},y_0-y_{-(t-1)}])$.

To aggregate spatial and temporal knowledge for each proposal, the concatenation of local geometry feature and the local dynamic feature is fed into a ResNet block followed by a global average pooling. The pooled feature is then concatenated with the history path feature, and serves as the proposal-local feature, noted as $f_i$ for the i-th temporal proposal.

As discussed before, the future trajectory of a pedestrian could be influenced by the surrounding pedestrians' behaviors. In order to model such interactions among pedestrians, we design an interaction layer which uses a graph to propagate information among objects, as shown in the middle part of Figure~\ref{fig:feature_extraction}. Specifically, we represent each temporal proposal as a graph node $i$; the embedding of node $i$ is noted as $f_i$, which is the corresponding proposal-local feature. The edge $v_{ij}$ represents the interaction score between node $i$ and node $j$. $v_{ij}$ is learned from $f_i$ and $f_j$, which can be represented as below.
\begin{equation*}
v_{ij}=\alpha([\phi_1(f_i);\phi_2(f_j)])
\end{equation*}
where $\alpha$ and $\phi$'s can be any learnable functions. In our implementation, we use fully-connected layer for $\alpha$ and $\phi$'s. 

Given the interaction scores among all pairs of nodes, we can gather the information for each node from the neighboring nodes. Specifically, the interaction embedding $g_i$ gathered for node $i$ is calculated as follows:
\begin{equation*}
g_{i}=\sum_j\frac{\exp{\{v_{ij}\}}}{V_i}\gamma([f_i;f_j])
\end{equation*}
where $V_i=\sum_j\exp{\{v_{ij}\}}$ is the normalization constant, and $\gamma$ is a mapping function (a fully-connected layer is adopted in our implementation). 

\subsubsection{Proposal classification and regression}
Given proposal-local features $f_i$ for each temporal proposals, two fully-connected layers are applied to do classification and regression respectively for the current frame. To be aligned with our intuitions, the proposal-local feature $f_i$ combined with the interaction feature $g_i$ is used to predict future frame boxes, by one fully-connected layer with $3t$ output channels where $t$ is the number of future frames to predict and $3$ stands for x coordinate, y coordinate and heading respectively. During the training, temporal proposals are assigned classification and regression targets with the same strategy discussed in Subsection~\ref{sec:trpn} and the objective is the weighted sum of classification loss, current frame regression loss and future frames regression loss similar to Equations~\ref{eq:6}-\ref{eq:9}. During inference, each proposal is predicted with a classification score and current/future boxes. Non-maximum suppression is applied on them based on the IoU between their current boxes, to remove redundancy.

\section{Experiment}
\begin{table*}[!tb]
\centering
\begin{tabular}{c|c|c||c|c|c|c|c|c|c}\hline
Model      & MF         & TS        & DE@1 $\downarrow$           & DE@2 $\downarrow$           & DE@3 $\downarrow$           & ADE $\downarrow$            & HR@1 $\uparrow$             & HR@2 $\uparrow$             & HR@3 $\uparrow$              \\\hline\hline
IntentNet  & \checkmark &           & \apmb{21.17}{0.02}          & \apmb{39.74}{0.07}          & \apmb{61.60}{0.12}          & \apmb{36.04}{0.12}          & \apmb{93.18}{0.03}          & \apmb{76.50}{0.08}          & \apmb{61.60}{0.12}           \\\hline
MF-FRCNN & \checkmark &\checkmark & \apmb{20.87}{0.08}          & \apmb{39.23}{0.14}          & \apmb{60.59}{0.22}          & \apmb{35.57}{0.13}          & \apmb{93.45}{0.05}          & \apmb{76.69}{0.18}          & \apmb{61.57}{0.21}           \\\hline
STINet     & \checkmark &\checkmark & \textbf{\apmb{19.63}{0.03}} & \textbf{\apmb{37.07}{0.08}} & \textbf{\apmb{57.60}{0.14}} & \textbf{\apmb{33.67}{0.07}} & \textbf{\apmb{94.36}{0.05}} & \textbf{\apmb{78.91}{0.06}} & \textbf{\apmb{64.43}{0.15}}  \\\hline
\end{tabular}
\caption{Trajectory prediction performance for different models on WOD. MF indicates whether the corresponding model takes multiple frames as input. TS indicates whether the model has a two-stage framework. $\uparrow$ and $\downarrow$ indicate the higher/lower numbers are better for the corresponding metric. DE and ADE are in centimeters. For models implemented by us, we train and evaluate the model for five times and compute the average and standard deviation shown around $\pm$ in the table.}\label{tbl:result_traj}
\end{table*}
\begin{table}[!tb]
\centering
\begin{tabular}{c|c|c||c}\hline
Model                                            & MF         & TS         & BEV AP $\uparrow$           \\\hline\hline
PointPillar~\cite{zhou2019end}                   &            &            & 68.57                       \\\hline
MVF~\cite{zhou2019end}                           &            &            & 74.38                       \\\hline
StarNet~\cite{ngiam2019starnet}                  &            &            & 72.50                       \\\hline
IntentNet~\cite{casas2018intentnet}\footnotemark & \checkmark &            & \apmb{79.43}{0.10}          \\\hline
MF-FRCNN                                       & \checkmark & \checkmark & \apmb{79.69}{0.19}          \\\hline
STINet                                           & \checkmark & \checkmark & \textbf{\apmb{80.73}{0.26}} \\\hline
\end{tabular}
\caption{Detection performance for different methods on WOD. MF indicates whether the corresponding model takes multiple frames as input. TS indicates whether the model has a two-stage framework. BEV AP is computed with an IoU threshold of 0.5. $\uparrow$ indicates the higher numbers are better for the corresponding metric.}\label{tbl:result_det}
\end{table}

\subsection{Experiment settings}
\noindent
\textbf{Dataset:} We conduct experiments on the Waymo Open Dataset (WOD)~\cite{waymo2019open} and the Lyft Dataset (Lyft)~\cite{lyft2019}. WOD contains lidar data from 5 sensors and labels for 1000 segments. Each segment contains roughly 200 frames and has a length of 20 seconds. Train and validation subsets have 798 and 202 segments respectively. To model the history and predict the future, we take 1 second history frames and 3 second future frames for each example and extract examples from the center 16 seconds (1s$\sim$17s) from each segment. Thus 126,437 train examples and 31,998 validation examples are extracted, and each of them contains history frames of 1 second and future frames of 3 seconds. We sample 6 frames including 5 history frames and the current frame, with $t_{\text{input}}=\{-1.0,-0.8,-0.6,-0.4,-0.2,0\}$, and the point clouds from those frames are fed into the network as inputs. In order to build richer voxel features while saving computation and memory, every two frames are combined by concatenating the voxelization output features thus we have three pillar features as discussed in Subsection~\ref{sec:backbone}. For the future prediction, we predict trajectory for 6 future frames with $t_{\text{future}}=\{0.5,1.0,1.5,2.0,2.5,3.0\}$. The range is 150m by 150m around the self-driving car, and we use a pillar size of 31.25cm by 31.25cm to generate pillar features of shape $480\times480$. Lyft contains lidar data from 1 sensor and labels for only 180 segments, with 140 and 40 segments for train and validation respectively. With the same settings, 14,840 and 4,240 examples are extracted for train and validation. Each example has 1-second history and 3-second future. We have $t_{\text{future}}=\{0.6,1.2,1.8,2.4,3.0\}$ for Lyft due to its 5Hz sampling rate.
\footnotetext{IntentNet without intent prediction head implemented by us.}

\noindent
\textbf{Evaluation metric:} The evaluation metric for detection is BEV AP (Bird-Eyes-View Average Precision) with the IoU threshold set to 0.5. Objects with fewer than 5 points are considered hard and are excluded during evaluation. For trajectory prediction, we employ the metrics used in~\cite{casas2018intentnet,hong2019rules}. For $t\in t_{\text{future}}$, we compute the DE@$t$ (Displacement Error) and the HR@$t$ (Hit Rate) with a displacement error threshold of 0.5m. We also compute the ADE (Average Displacement Error) which equals to $\frac{1}{|t_{\text{future}}|}\sum_{t\in t_{\text{future}}}\text{DE@}t$.

\noindent
\textbf{Implementation:} Our models are implemented in TensorFlow and we train the model with Adam optimizer on TPUv3 for 140k and 70k iterations for Waymo Open Dataset and Lyft Dataset respectively. The learning rate is $4\times10^{-4}$ and batch size is 1 per TPU. We use 32 TPU cores together for the training, thus the effective batch size is 32. We also implement IntentNet~\cite{casas2018intentnet} and Faster-RCNN~\cite{ren2015faster} in TensorFlow as the baselines, which are noted as ``IntentNet'' and ``MF-FRCNN''. Our implemented IntentNet (1) takes multiple frames as input and share the same backbone net as STINet; (2) removes the intent classification part, and only regresses a future trajectory. MF-FRCNN refers to a Faster-RCNN~\cite{ren2015faster} model with several changes: (1) It uses the same backbone net as STINet, please refer to Section \ref{sec:backbone}; (2) for each object proposal, in addition to the bounding box, we also regress future trajectories and headings. Note that the difference between proposals from MF-FRCNN and our method is that MF-FRCNN only predicts the current box of objects, while our method exploits a novel Temporal RPN which also generates the corresponding history boxes associated to each current box.

\subsection{Results on Waymo Open Dataset}\label{sec:result_main}

The main results on Waymo Open Dataset of pedestrian detection and trajectory prediction are summarized in Table~\ref{tbl:result_det} and Table~\ref{tbl:result_traj}. For detection we compare our proposed method (in the last row) with the current state-of-the-art detectors~\cite{ngiam2019starnet,zhou2019end} and our method surpasses the off-the-shelf baselines by a very large margin, improving the BEV AP from 74.38 to 80.73. To avoid the effects from multi-frame inputs and different implementation details, we also compare with our implementation of IntentNet and multi-frame Faster RCNN \cite{ren2015faster}, which are noted as ``IntentNet'' and ``MF-FRCNN'' in Table~\ref{tbl:result_det}. Our proposed method outperforms all baselines and it confirms the effectiveness of our T-RPN and the \ifsti STI \fi modeling of proposals.

In Table~\ref{tbl:result_traj} we compare the trajectory prediction performance between our proposed method, IntentNet and MF-FRCNN. Our proposed method surpasses all competitors by a large margin, and the improvement is larger than the improvement on detection. It aligns with our intuition since T-RPN and \ifsti STI \fi modeling are designed to better model objects' movement and more useful to forecast their trajectory.\par
\begin{table}[!tb]
\centering
\small
\begin{tabular}{c||c|c|c|c|c}\hline
Model         & \small{0$\sim$2.5} & \small{2.5$\sim$5} & \small{5$\sim$7.5} & \small{7.5$\sim$10} & \small{10$\sim$ $\infty$} \\\hline\hline
MF-FRCNN    & 63.07              & 90.44              & 93.27              & 88.00               & 77.15                     \\\hline
STINet        & \textbf{64.23}     & \textbf{91.15}     & \textbf{94.46}     & \textbf{88.97}      & \textbf{80.50}            \\\hline
$\Delta\%$    & 1.8\%              & 0.8\%              & 1.3\%              & 1.1\%               & 4.3\%                     \\\hline
\end{tabular}
\caption{Bird-eyes-view average precision (BEV-AP) breakdown comparison of MF-FRCNN and STINet on WOD. Objects are split into five bins base on the future trajectory length with a bin size of 2.5m. Last row is the relative improvement of STINet.}\label{tbl:distance_breakdown_det}
\end{table}
For a detailed comparison of \ifsti STINet \fi and MF-FRCNN, we evaluate the detection and trajectory prediction by breaking down the objects into five bins based on the future trajectory length in 3s. The five bins are 0$\sim$2.5m, 2.5$\sim$5m, 5$\sim$7.5m, 7.5$\sim$10m and 10m$\sim$ $\infty$ respectively. We report BEV AP, ADE and the relative improvement in Table~\ref{tbl:distance_breakdown_det} and~\ref{tbl:distance_breakdown_tra}. The \ifsti STINet \fi is consistently better than MF-FRCNN for both tasks. For trajectory prediction on objects moving more than 5m, the relative improvements are significant and consistently more than 10\%. It confirms that the proposed method can leverage the details of history information and provide much better trajectory predictions, especially for pedestrians with a larger movement.

\begin{table}[!tb]
\centering
\small
\begin{tabular}{c||c|c|c|c|c}\hline
Model         & \small{0$\sim$2.5} & \small{2.5$\sim$5} & \small{5$\sim$7.5} & \small{7.5$\sim$10} & \small{10$\sim$ $\infty$} \\\hline\hline
MF-FRCNN    & 26.90              & 37.56              & 46.39              & 104.60              & 173.50                    \\\hline
STINet        & \textbf{26.73}     & \textbf{35.42}     & \textbf{41.18}     & \textbf{89.74}      & \textbf{137.17}           \\\hline
$\Delta\%$    & 0.6\%              & 6.0\%              & 11.2\%             & 14.2\%              & 20.9\%                    \\\hline
\end{tabular}
\caption{Average displacement error (ADE, in centimeters) breakdown comparison of MF-FRCNN and STINet on WOD. Objects are split into five bins base on the future trajectory length with a bin size of 2.5m. Last row is the relative improvement of STINet.}\label{tbl:distance_breakdown_tra}
\end{table}

\subsection{Results on Lyft Dataset}
The detection and trajectory prediction results on the Lyft Dataset are summarized in Table~\ref{tbl:result_lyft}. The performances on both tasks are improved largely and the results confirm the effectiveness of proposed method a small-scale dataset.

\begin{table}[!tb]
\centering
\small
\begin{tabular}{c||c|c|c|c}\hline
Model      & BEV AP $\uparrow$ & DE@3 $\downarrow$ & ADE $\downarrow$ & HR@3 $\uparrow$ \\\hline\hline
MF-FRCNN & 33.90             & 82.61             & 51.11            & 49.74           \\\hline
STINet     & \textbf{37.15}    & \textbf{76.17}    & \textbf{46.09}   & \textbf{50.73}  \\\hline
\end{tabular}
\caption{Detection and trajectory prediction performance on Lyft.}\label{tbl:result_lyft}
\end{table}

\begin{table}[!tb]
\centering
\small
\begin{tabular}{c|c||c|c|c|c}\hline
LG         & LD         & BEV AP $\uparrow$ &  DE@3 $\downarrow$ & ADE $\downarrow$ & HR@3 $\uparrow$ \\\hline\hline
\checkmark &            & 80.38             & 64.15              & 37.67            & \text{58.46}    \\\hline
           & \checkmark & 79.69             & 59.71              & 34.96            & \text{62.22}    \\\hline
\checkmark & \checkmark & \textbf{80.53}    & \textbf{58.95}     & \textbf{34.49}   & \textbf{62.99}  \\\hline
\end{tabular}
\caption{Ablation studies on local geometry and local dynamic features (noted as LG and LD in the table respectively). All entries are trained without History Path and Interactive features.}\label{tbl:ablation_local_global}
\end{table}

\begin{table}[!tb]
\centering
\begin{tabular}{c|c||c|c|c}\hline
L+G        & Path       & DE@3 $\downarrow$ & ADE $\downarrow$ & HR@3 $\uparrow$ \\\hline\hline
\checkmark &            & 58.95             & 34.49            & 62.99           \\\hline
\checkmark & \checkmark & \textbf{58.04}    & \textbf{33.92}   & \textbf{63.87}  \\\hline
$\dagger$  & \checkmark & 67.80             & 39.86            & 52.25           \\\hline
\end{tabular}
\caption{Ablation studies on history path feature. $\dagger$ indicates the corresponding feature is used only for detection and ignored while prediction the trajectory.}\label{tbl:ablation_history_path}\vspace{-0.3cm}
\end{table}

\begin{table}[!tb]
\centering
\begin{tabular}{c||c||c|c|c}\hline
Breakdown                & I          & DE@3 $\downarrow$ & ADE $\downarrow$ & HR@3 $\uparrow$ \\\hline\hline
\multirow{2}{*}{All}     &            & 58.04             & 33.92            & 63.87           \\\cline{2-5}
                         & \checkmark & \textbf{57.60}    & \textbf{33.67}   & \textbf{64.43}  \\\hline\hline
\multirow{2}{*}{Group}   &            & 49.67             & 30.85            & 64.87           \\\cline{2-5}
                         & \checkmark & \textbf{48.89}    & \textbf{30.40}   & \textbf{65.55}  \\\hline
\end{tabular}
\caption{Ablation studies on interaction features. `I' indicates whether the proposal interaction modeling is adopted. ``All" and ``Group" correspond to evaluation on all pedestrians and pedestrians belonging to a group with at least 5 pedestrians respectively.}\label{tbl:ablation_interaction}
\end{table}

\subsection{Ablation studies}\label{sec:result_ablation}
In this section we conduct ablation experiments to analyze the contribution of each component and compare our model with potential alternative methods on the Waymo Open Dataset. The results are summarized below. For clarity, we only show DE@3, ADE and HR@3 for trajectory prediction. The other metrics have the same tendency.

\noindent
\textbf{Effect of local geometry and local dynamic features:} We conduct experiments to analyze the effect of local geometry and local dynamic features, summarized in Table~\ref{tbl:ablation_local_global}. The local geometry feature is good at detection and the local dynamic feature is good at trajectory prediction. Geometry feature itself does not work well for trajectory prediction since it ignores dynamics for better detection. By combining both of the features, the benefits in detection and trajectory prediction can be obtained simultaneously.

\noindent
\textbf{Effect of history path:} Although objects' geometry and movement are already represented by local geometry dynamic features, taking history path as an extra feature can give another performance gain by improving the DE@3 from 58.95 to 58.04 and the HR@3 from 62.99 to 63.87 (as shown in the first two row of Table~\ref{tbl:ablation_history_path}). This suggests the history path, as the easiest and most direct representation of objects' movement, can still help based on the rich representations. However history path itself is far from enough to give accurate trajectory prediction, suggested by the poor performance in the last row of Table~\ref{tbl:ablation_history_path}.

\noindent
\textbf{Effect of proposal interaction modeling:} 
To demonstrate the effectiveness of the proposed pedestrian interaction modeling, we measure the performance for all pedestrians as well as pedestrians in a group. Specifically, we design a heuristic rule (based on locations and speeds) to discover pedestrian groups and assign each pedestrian a group label on the evaluation set. The details about the grouping algorithm can be found in supplementary. We evaluate the trajectory prediction performance on all pedestrians and the pedestrians belonging to a group with at least 5 pedestrians, shown in Table~\ref{tbl:ablation_interaction}. The interaction modeling improves trajectory prediction performance on ``all pedestrians" and achieve a larger boost for pedestrians that belong to groups (DE@3 improved from 49.67 to 48.89 by 1.6\%).
\subsection{Model inference speed}
We measure the inference speed of our proposed model as well as baseline models on context range of 100m by 100m as well as 150m by 150m. All models are implemented in TensorFlow and the inference is executed on a single nVIDIA Tesla V100 GPU. For the context range of 100m by 100m, IntentNet, MF-FRCNN and STINet have inference time of 60.9, 69.4 and 74.6ms respectively. Both two-stage models (MF-FRCNN and STINet) are slower than the single-stage model, and STINet is slightly slower than MF-FRCNN. However, all three models can achieve a real-time inference speed higher than 10Hz. For the maximum range of Waymo Open Dataset, \ie, 150m by 150m, three models have inference time of 122.9, 132.1 and 144.7ms respectively.
\subsection{Qualitative results}
\begin{figure}
    \centering
    \includegraphics[width=\linewidth]{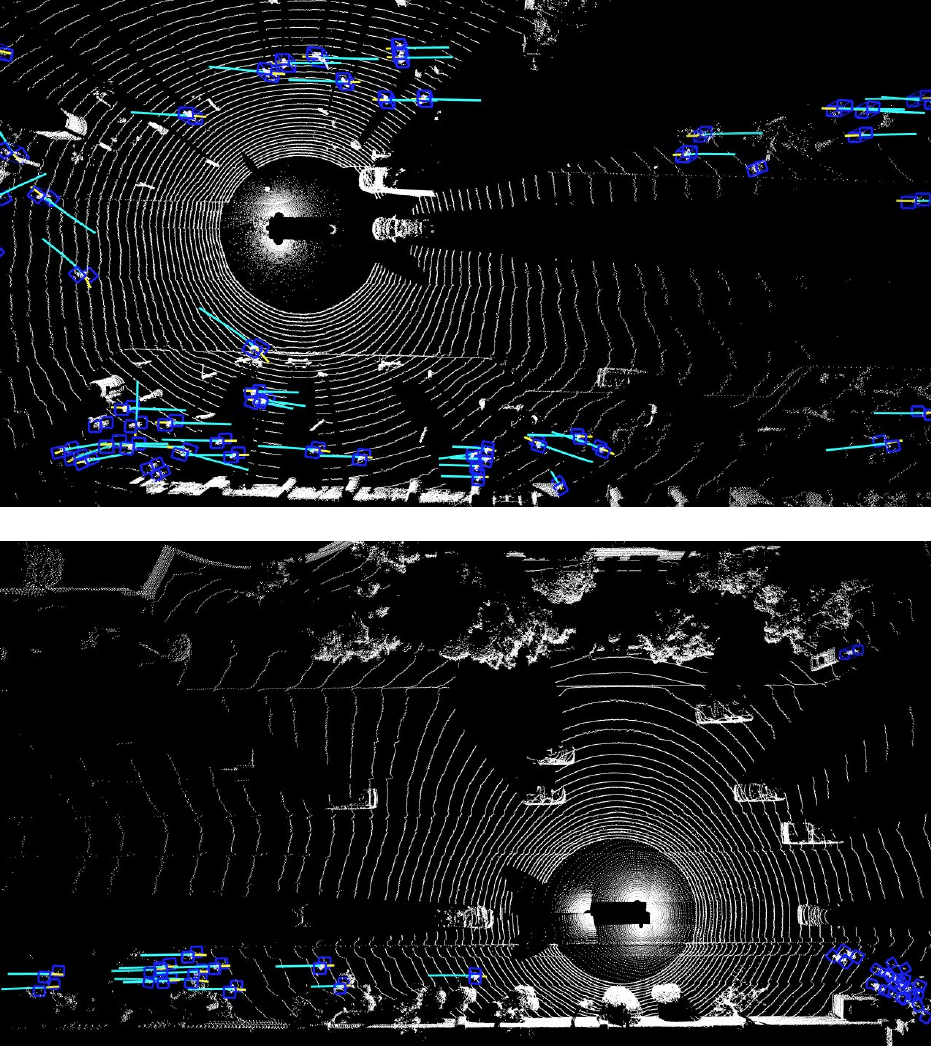}
    \caption{Qualitative examples of STINet. The blue box are detected pedestrians. The cyan and yellow lines are predicted future and history trajectories of STINet respectively.}
    \label{fig:qualitative}
\end{figure}
\begin{figure}
    \centering
    \includegraphics[width=\linewidth]{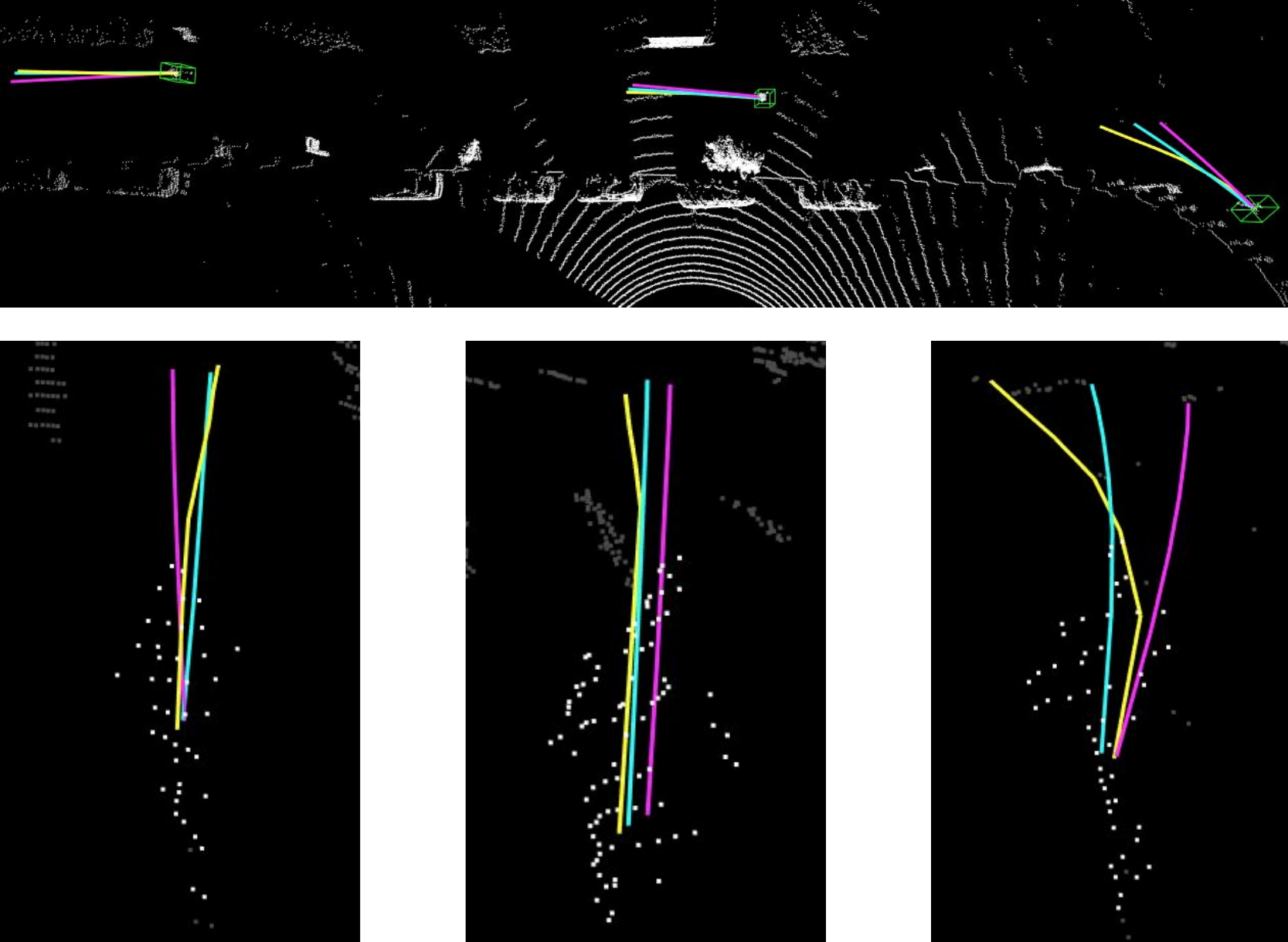}
    \caption{Comparison between MF-FRCNN and STINet. The yellow line is the ground-truth future trajectory for pedestrians. The pink and cyan lines are the predicted future trajectory from MF-FRCNN and STINet respectively. It is clear that our proposed method gives a much better prediction compared with the baseline, for all three pedestrians. Upper: the overview of three pedestrians. Lower: zoom-in visualization for three pedestrians.}
    \label{fig:comparison}
\end{figure}
The visualization for the predictions of \ifsti STINet \fi is shown in Figure~\ref{fig:qualitative}. The blue boxes are the detected pedestrians. The cyan and yellow lines are the predicted future and history trajectory for each detected pedestrian respectively. We show two scenarios where the SDC is stationary in the upper sub-figure and the SDC is moving fast in the lower sub-figure. It demonstrates that our model detects and predicts very accurately in both cases.

Figure~\ref{fig:comparison} shows a detailed comparison between \ifsti STINet \fi and MF-FRCNN against the ground-truth for trajectory prediction. Green boxes are the ground-truth boxes. Yellow, pink and cyan lines are the ground-truth future trajectory as well as the predicted future trajectories from MF-FRCNN and \ifsti STINet \fi respectively. For the left two pedestrians who are walking in a straight line, both MF-FRCNN and \ifsti STINet \fi predict future trajectory reasonably well but the MF-FRCNN still has a small error compared with the ground-truth; for the right-most pedestrian who is making a slight left turn, MF-FRCNN fails to capture the details of its movement and gives an unsatisfactory prediction, while \ifsti STINet \fi gives a much better trajectory prediction.

\section{Conclusion}
In this paper, we propose \ifsti STINet \fi to perform joint detection and trajectory prediction with raw lidar point clouds as the input. We propose to build temporal proposals with pedestrians' both current and past boxes and learn a rich representation for each temporal proposal, with local geometry, dynamic movement, history path and interaction features. We show that by explicitly modeling the \ifsti spatio-temporal-interaction \fi features, both detection and trajectory prediction quality can be drastically improved compared with single-stage and two-stage baselines. This also makes us to re-think the importance of introducing second-stage and proposals, especially for the joint detection and trajectory prediction task. Comprehensive experiments and comparisons with baselines and state-of-the-arts confirm the effectiveness of our proposed method, and our method significantly improves the prediction quality while still achieves the real-time inference speed which makes our model practical to be used in real-world applications. Combining camera/map data and utilizing longer history with LSTMs could be investigated to further improve the prediction and we will explore them in future work.
{\small
\bibliographystyle{ieee_fullname}
\bibliography{egbib}
}

\end{document}